\documentclass[11pt,a4paper]{article}
\usepackage[hyperref]{acl2020}
\usepackage{times}
\usepackage{latexsym}
\usepackage[linesnumbered,algoruled,boxed,noend]{algorithm2e}
\usepackage{amsmath, amssymb, bm}
\usepackage{url} 
\SetKwInOut{Parameter}{parameter}
\usepackage{enumitem}
\usepackage{mathtools}
\usepackage{cleveref}
\usepackage{multirow}
\usepackage{hhline}
\usepackage{subcaption}
\usepackage[export]{adjustbox}
\usepackage{booktabs,array}
\newcolumntype{P}[1]{>{\centering\arraybackslash}p{#1}}
\usepackage{booktabs,makecell,tabularx} 

\let\oldnl\nl
\newcommand{\nonl}{\renewcommand{\nl}{\let\nl\oldnl}}

\newcommand{\secref}[1]{\S\ref{#1}}
\newcommand{\figref}[1]{Fig. \ref{#1}}

\aclfinalcopy 

\title{LEAN-LIFE: A Label-Efficient Annotation Framework \\Towards Learning from Explanation}

\author{
Dong-Ho Lee\textsuperscript{1}\thanks{~~Both authors contributed equally.}~~~
Rahul Khanna\textsuperscript{1}$^*$~~
Bill Yuchen Lin\textsuperscript{1}~~
Jamin Chen\textsuperscript{1}~~
Seyeon Lee\textsuperscript{1}~~\\
{\bf Qinyuan Ye}\textsuperscript{1}~~
{\bf Elizabeth Boschee}\textsuperscript{2}~~
{\bf Leonardo Neves}\textsuperscript{3}~~
{\bf Xiang Ren}\textsuperscript{1,2}
\\
\textsuperscript{1}Department of Computer Science, University of Southern California\\
\textsuperscript{2}Information Science Institute, University of Southern California\quad\textsuperscript{3}Snap Inc.\\
{\small{\texttt{\{dongho.lee,rahulkha,yuchen.lin,jaminche,seyeonle,qinyuany\}@usc.edu}}},\\
{\small{\texttt{boschee@isi.edu},
\texttt{lneves@snap.com},
\texttt{xiangren@usc.edu}}}
}

\date{}

\begin{document}
\maketitle
\begin{abstract}


Successfully training a deep neural network demands a huge corpus of labeled data. However, each label only provides limited information to learn from and collecting the requisite number of labels involves massive human effort.
In this work, we introduce \texttt{LEAN-LIFE}\footnote{The source code is publicly available at~\url{http://inklab.usc.edu/leanlife/}.}, a web-based, \textbf{L}abel-\textbf{E}fficient \textbf{A}nnotatio\textbf{N} framework for sequence labeling and classification tasks, with an easy-to-use UI that not only allows an annotator to provide the needed labels for a task, but also enables \textbf{L}earn\textbf{I}ng \textbf{F}rom \textbf{E}xplanations for each labeling decision.
Such explanations enable us to generate useful additional labeled data from unlabeled instances, bolstering the pool of available training data. On three popular NLP tasks (named entity recognition, relation extraction, sentiment analysis), we find that using this enhanced supervision allows our models to surpass competitive baseline F1 scores by more than 5-10 percentage points, while using 2X times fewer labeled instances. 
Our framework is the first to utilize this enhanced supervision technique and does so for three important tasks––thus providing improved annotation recommendations to users and an ability to build datasets of \textit{(data, label, explanation)} triples instead of the regular \textit{(data, label)} pair.

\end{abstract}

\section{Introduction}
Deep neural networks have achieved state-of-the-art performance on a wide range of sequence labeling and classification tasks such as named entity recognition (NER)~\cite{DBLP:conf/naacl/LampleBSKD16,DBLP:conf/acl/MaH16}, relation extraction (RE)~\cite{zeng2015distant, zhang2017position, Ye2019LookingBL}, and sentiment analysis (SA)~\cite{wang2016attention}.
However, they only yield such performance levels in supervised learning scenarios, and in particular when human-annotated data is abundant. As we seek to apply NLP models to larger variety of domains, such as product reviews~\cite{Luo2018ExtRAEP}, social media messages~\cite{Lin2017MultichannelBM}, while reducing human annotation efforts, better annotation frameworks with label-efficient learning techniques are crucial to our progress.

\begin{figure}[t]
	\centering
	\setlength\abovecaptionskip{-0.01\baselineskip}
	\setlength\belowcaptionskip{-0.2\baselineskip}
	\vspace*{-.1in}
	\hspace*{-.2in}
	\includegraphics[width=1.1\linewidth]{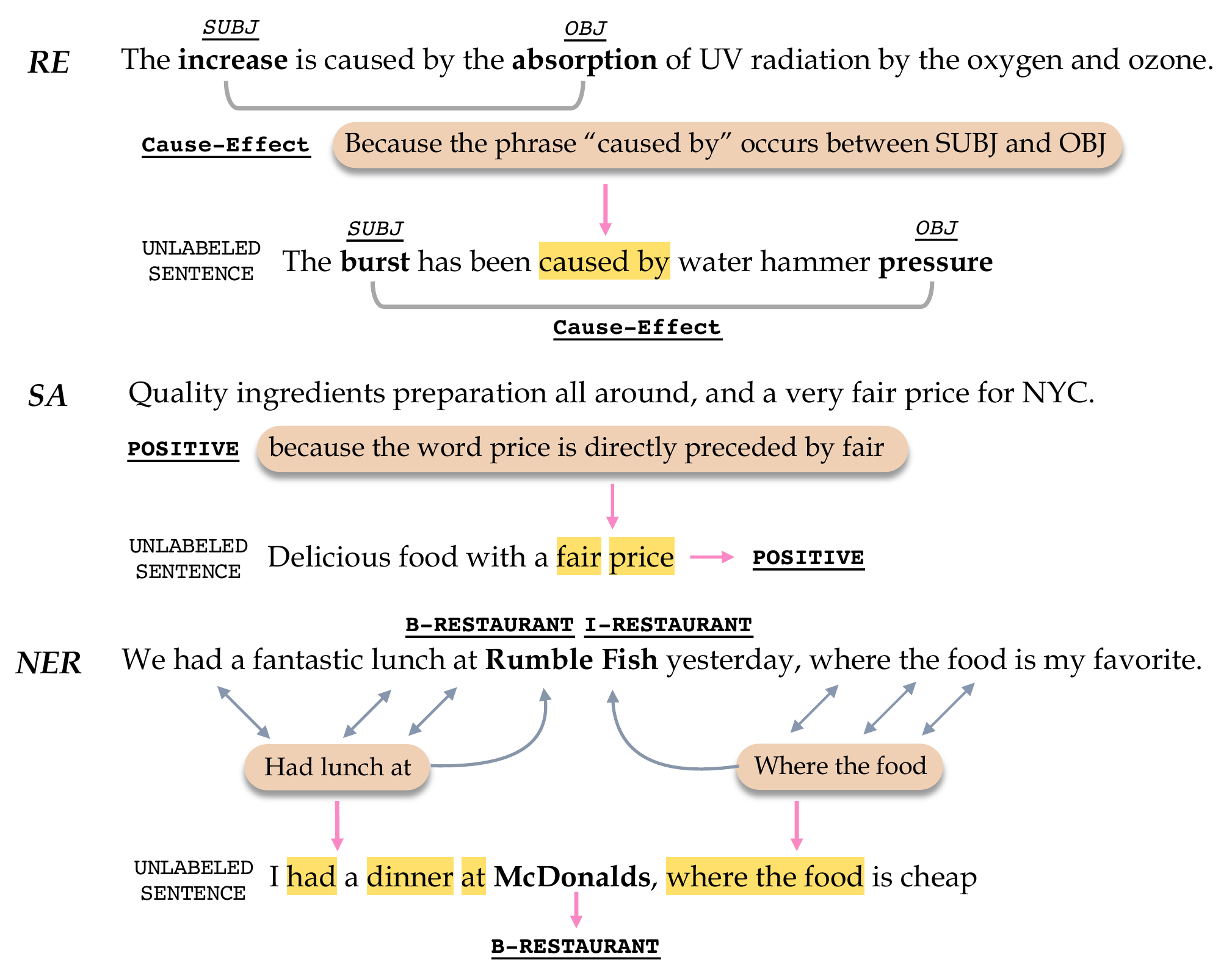}
	\caption{\textbf{Leveraging Labeling Explanations:}
	\textbf{1) RE}: the explanation \textit{``the phrase `caused by' occurs between SUBJ and OBJ"} can aid in weakly labeling unlabeled instances like \textit{``The burst has been caused by water hammer pressure"} with the label \textit{``cause-effect"};
	\textbf{2) NER}: Trigger spans near the labeled restaurant such as \textit{``had lunch at"} and \textit{``where the food"} can aid in weakly labeling unlabeled instances like \textit{``I had a dinner at McDonalds, where the food is cheap"}. }
	\vspace{-10pt}
	\label{fig:exp}
\end{figure}

Annotation frameworks have been explored by several previous works~\cite{stenetorp2012brat, bontcheva2014gate, morton2003wordfreak, de2016web, yang2017yedda}. These existing open-source sequence annotation tools mainly focus on optimizing user-friendly user interfaces, such as providing shortcut key functionality to allow for faster tagging.
The frameworks also attempt to provide annotation recommendation to reduce human annotation efforts.
However, these recommendations are provided by a pre-trained model or via dictionary look-ups. 
This methodology of providing recommendations often proves to be unhelpful when little annotated data exists for pre-training, as is usually the case for natural language tasks being applied to domain-specific or user-provided corpora.

To resolve this issue, AlpacaTag, an annotation framework for sequence labeling ~\cite{lin2019alpacatag} attempts to provide annotation recommendations from a learned sequence labeling model that is incrementally updated by batches of incoming human annotations. 
Its model training follows an active learning strategy~\cite{shen2017deep}, which is shown to be a label-efficient, thus it attempts to minimize human annotation efforts. AlpacaTag selects the most informative batches of documents for humans to annotate and thus achieves a more cost-effective way of using human efforts. 
While active learning allows the model to achieve higher performance earlier in the learning process, model performance could be improved if additional supervision existed.
It is imperative that provided annotation recommendations be as accurate as possible, as inaccurate annotation recommendations from the framework can push users towards generating noisy data, hindering instead of aiding the model training process.

Our effort to prevent this problem is centered around allowing annotators to provide additional supervision by capturing labeling explanations, while still taking advantage of the cost-effectiveness of active learning. Specifically, as shown in ~\figref{fig:exp}, we allow annotators to provide explanations for their decisions in natural language or by selecting \textit{triggers}––nearby phrases that provide helpful context for their decisions. 
These enhanced annotations allow for model training over both user-provided labels, as well as weakly labeled data created by parsing explanations into high precision labeling rules.
We therefore make attempts to ameliorate the erroneous recommendation problem by a performance-boosting training strategy that incorporates both labeled and unlabeled data.

Our work is also similar to recent attempts that exploit explanations for an improved training process~\cite{srivastava2017joint,hancock2018training,wenxuan2020,qin2020learning}, but with two main differences. First, we embed this improved training process in a practical application and second, we design task specific architectures to incorporate the now captured explanations into training. 

To the best of our knowledge, there is no existing open-source, easy-to-use, recommendation-providing, online-learning annotation framework that can also capture explanations. \texttt{LEAN-LIFE} is the first framework to capture and leverage explanations for improved model training and performance, while still inheriting the advantages of existing tools. 
We summarize our contributions as:

\noindent
\textbullet\ \hspace{0.2mm} \textbf{Improved Model Training:}
Our recommendation models use a performance improving training process that leverages explanations to weakly label unlabeled instances. Our models improve on competitive baseline F-1 scores by more than 5-10 percentage points, while using 2X less data.

\noindent
\textbullet\ \hspace{0.2mm} \textbf{Multiple Supported Tasks:}
Our framework supports both sequence labeling (as in NER) and sequence classification (as in RE, SA).

\noindent
\textbullet\ \hspace{0.2mm} \textbf{Explanation Dataset Creation:} 
We make it easy to build a new type of dataset, one that consists of triples of: text, labels and labeling explanations. The exporting of this captured data is available in two common data formats, CSV and JSON.

\section{System Overview}
\begin{figure}[t]
	\centering
	\setlength\belowcaptionskip{-0.2\baselineskip}
	\vspace*{-.1in}
	\hspace*{-.2in}
	\includegraphics[width=\linewidth]{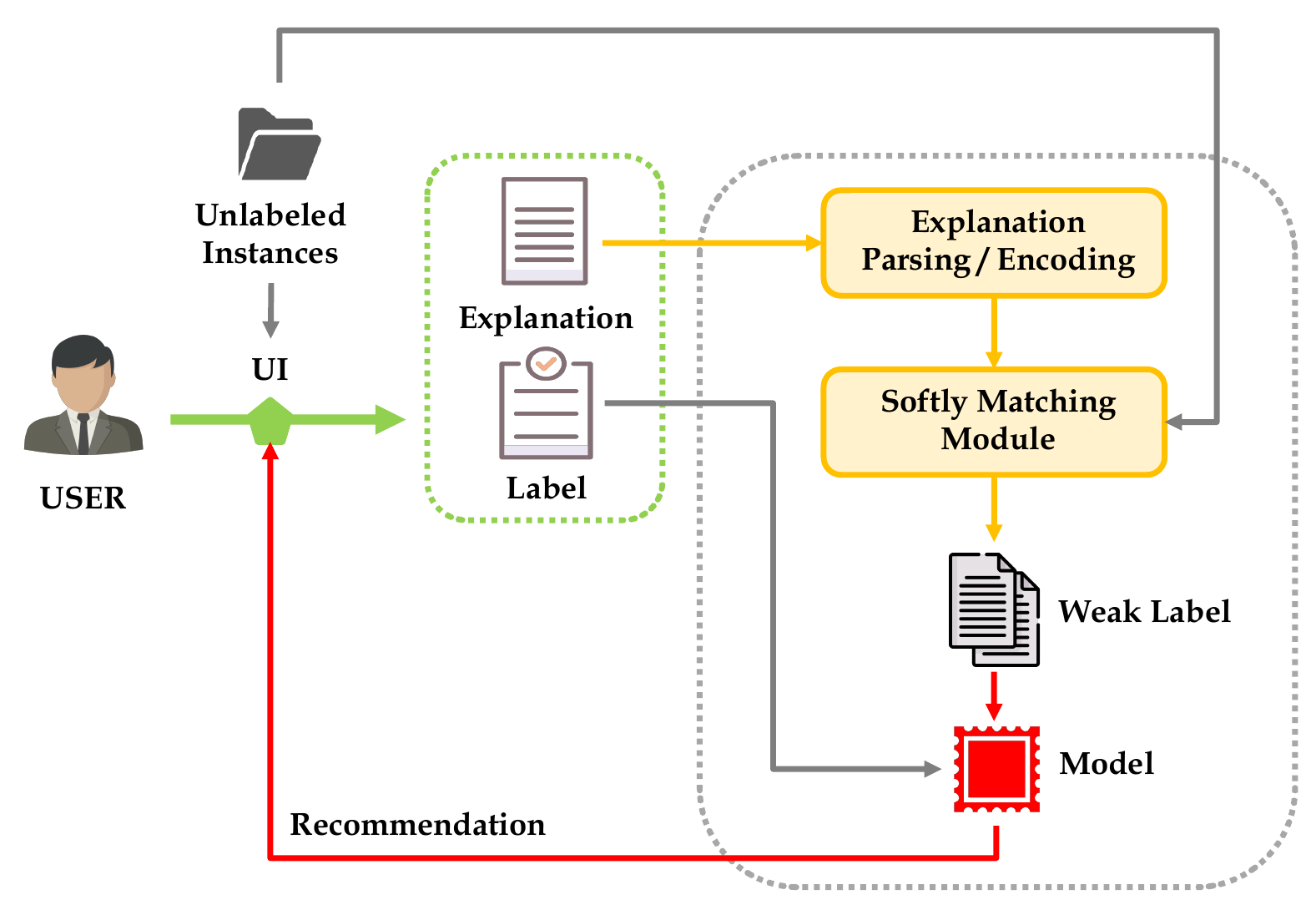}
	\caption{\textbf{System Architecture}.}
	\vspace{-15pt}
	\label{fig:overview}
\end{figure}

As shown in ~\figref{fig:overview}, our framework consists of two main components, a user-friendly web-UI that can capture labels and explanations for labeling decisions, and a weak supervision framework that parses explanations for the creation of weakly labeled data. The framework then uses this weakly labeled data in conjunction with user-provided labels to train models for improved annotation recommendations.
Our UI shows annotators unlabeled instances (can be sampled using active learning), along with annotation recommendations in an effort to reduce annotation costs.
We use PyTorch to build our models and implement an API for communication between the web-UI and our weak supervision framework.
The learned parameters of our framework are updated in an online fashion, thus improving in near real time.
We will first touch on the annotation UI (\secref{sec:ui}) and then go into our weak supervision framework (\secref{sec:model}).
\section{UI for Capturing Human Explanation}
\label{sec:ui}
The emphasis of our front-end design is to simplify the capture of both label and explanation for each labeling decision, while  reducing annotation effort via accessible annotation recommendation.
Our framework supports two forms of explanations, \textit{Triggers} and \textit{Natural Language}.
A Trigger is a group of words in the sentence being annotated that aided the annotator's labeling decision, while Natural Language is a written explanation of the labeling decision.
This  section  presents first the UI for capturing triggers (\secref{sec:trigger}) and then the UI for capturing natural language explanations (\secref{sec:nl}).
\subsection{Capturing Triggers}
\label{sec:trigger}
\begin{figure}[!t]
	\centering
	\setlength\belowcaptionskip{-1\baselineskip}
	\includegraphics[width=\linewidth]{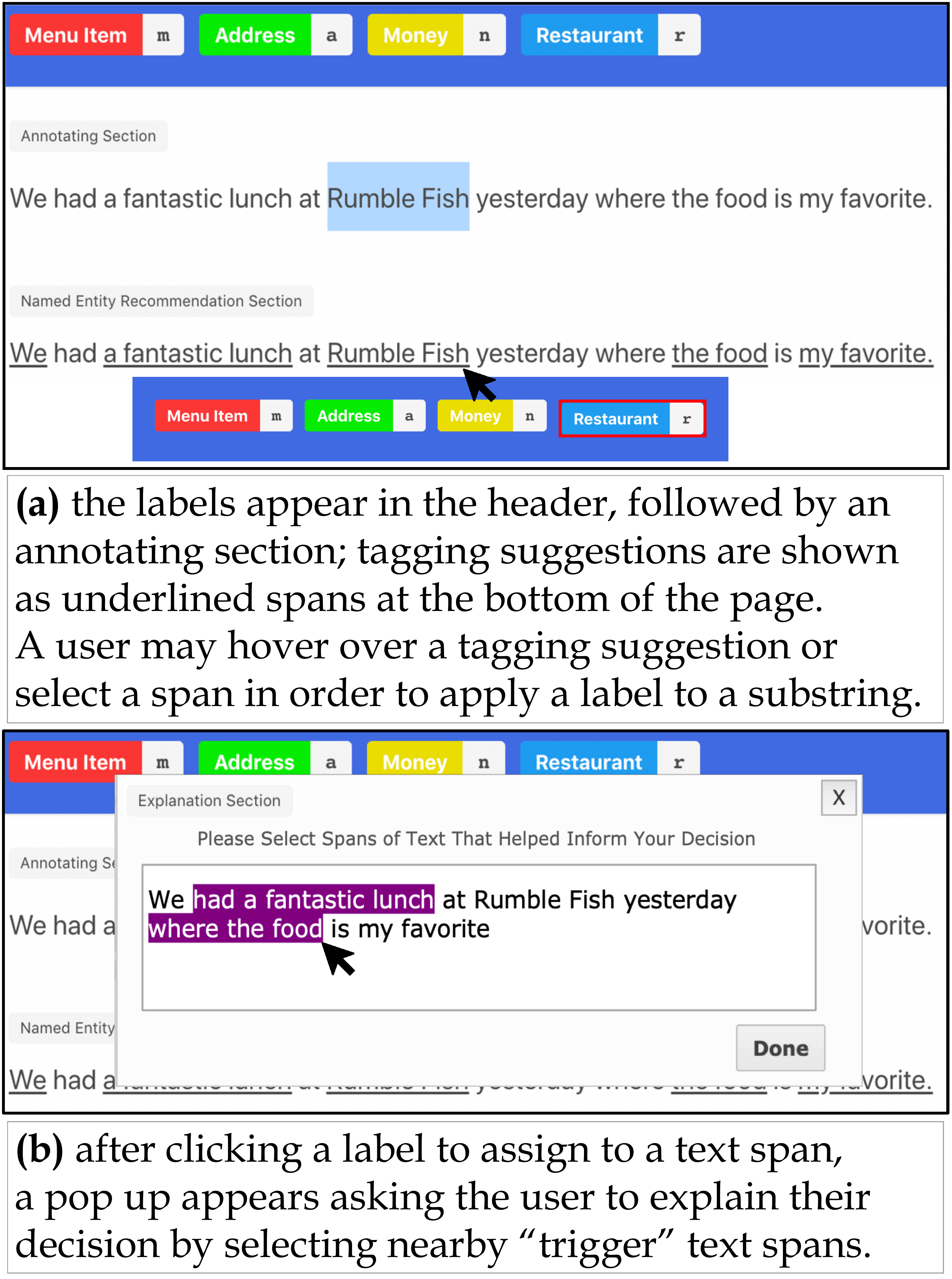}
	\caption{The workflow for annotators to annotate a NE label and trigger span. (“Rumble Fish” as Restaurant).}
	\label{fig:nerui}
\end{figure}

\figref{fig:nerui} illustrates how our framework can capture both a named entity (NE) label and triggers for the sentence ``We had a fantastic lunch at Rumble Fish yesterday where the food is my favorite''. 
The user is first presented with a piece of text to annotate (Annotating Section), the available labels that may be applied to sub-sequences (spans) of text (in the blue header) and 
recommendations of what spans of text should be considered as NE mentions (Named Entity Recommendation Section). 
The user may choose to select a span of text to label, or they may click on one of the recommended spans below (Fig. 2a). 
If the user clicks on a recommended span, a small pop-up displaying the available labels appear with the recommended label circled in red (Fig. 2a). 
Once the user selects a label for a span of text by either clicking on the desired label button or via a predefined shortcut key (ex: for Restaurant the shortcut key is \textbf{r}), a pop-up appears (Fig. 2b), asking the user to select helpful spans (triggers) from the text that provide useful context in deciding the label for the NEM––multiple triggers may be selected. 
The user may cancel their decision to label a span of text with a label by clicking the \textit{x} button in the pop-up, but if the user wants to proceed and has selected at least one trigger, they finish the labeling by hitting done. Then, their label is visualized in the Annotating Section by highlighting the NEM.

\subsection{Capturing Natural Language}
\label{sec:nl}
\begin{figure}[!t]
	\centering
	\setlength\belowcaptionskip{-1\baselineskip}
	\includegraphics[width=\linewidth]{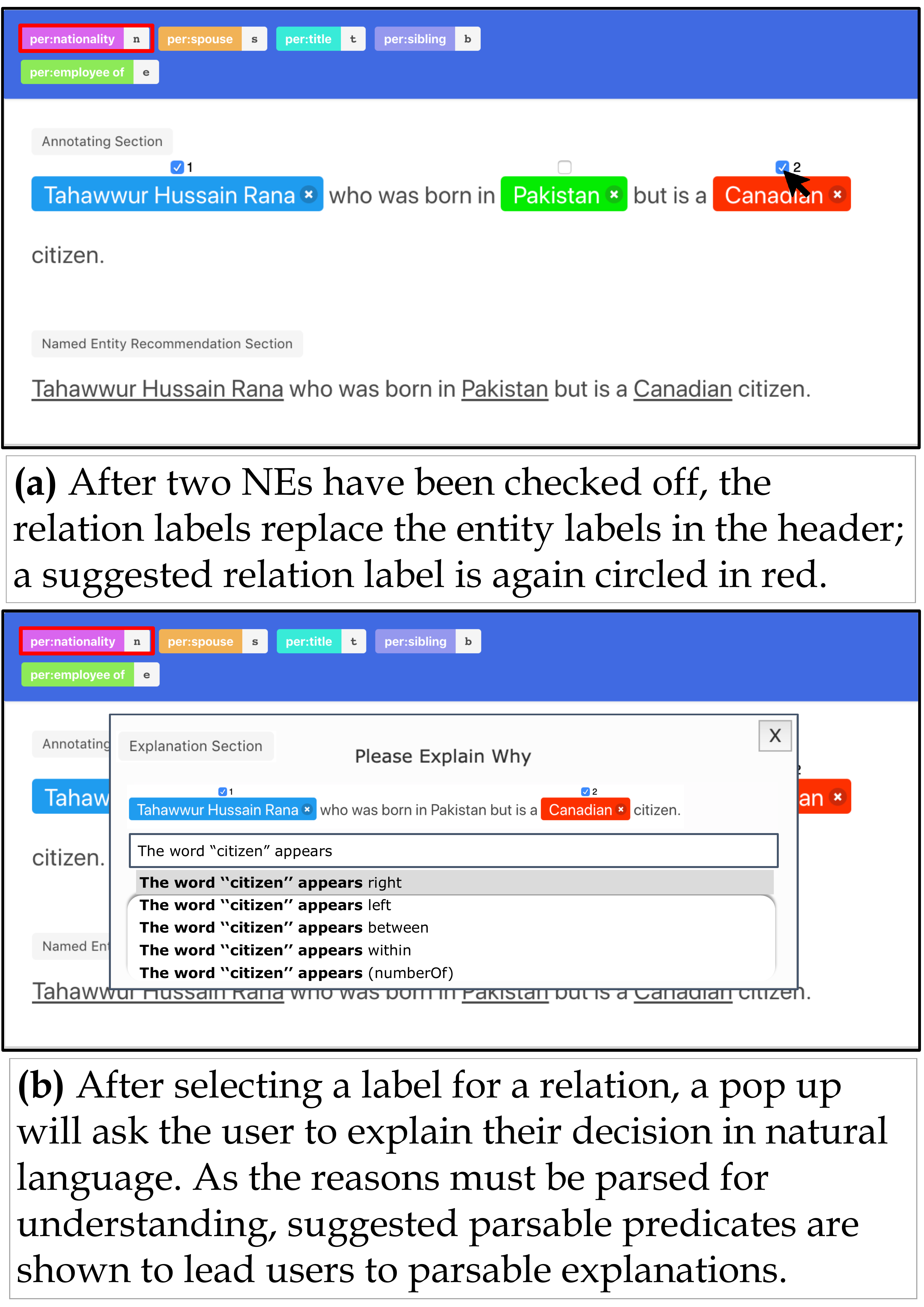}
	\caption{The workflow for annotators to annotate a relation label and NL explanation. (“per:nationality” as relation label between ``Tahawwur Hussain Rana" and ``Canadian").}
	\label{fig:reui}
\end{figure}

~\figref{fig:reui} illustrates how for the sentence ``Tahawwur Hussain Rana who was born in Pakistan but is a Canadian citizen" our framework can capture both a relation label between NEs and the subsequent natural language explanation.
First, the user is tasked to find the NEs in the sentence. 
After labeling at least two non-consecutive spans of text as NEs, the user may check off the boxes that appear above the labeled NEs. Once two boxes have been checked off, the labels in the blue header are replaced with the labels for relations. The click-order of the checked boxes is displayed and is considered the order of the relation. 
Also, we display a recommend label to the user in the header section with a circle (Fig. 2a).
After clicking on a label, a pop-up appears asking the user to indicate semantic and syntactic reasons as to why the labeling decision is true. 
Since the natural language explanations are assumed to be made up of predefined predicates, as the user types we incrementally provide predicates to aid the construction of an explanation (Fig. 2b).
In this way, we nudge users towards writing explanations the semantic parser is able to break down, allowing our framework to extract a useful logical form from the explanation.
\section{\texttt{LEAN-LIFE} Framework}
\label{sec:model}
Our Weak Supervision Framework is composed of two main components, a weak labeling module that parses explanations to create labeling rules and a downstream model.
The framework parses user-provided explanations to generate weakly labeled data and then trains the appropriate downstream model with this augmented training data. Our weak labeling module supports both explanation formats provided to the annotator in the UI––triggers and natural language. This  section  first  introduces how the module utilizes triggers (\secref{sec:modeltrigger}) and then presents how the module deals with natural language(\secref{sec:modelnl}).

\subsection{Input: Trigger}
\label{sec:modeltrigger}
When a trigger is inputted into the system, we generate weak labels for our training data via soft-matching between trigger representations and unlabeled sentences~\cite{TriggerNER2020}.
Each sentence may contain one or more triggers, but each trigger is associated with only one label.
Our framework jointly learns a mapping between triggers and their label using a linear layer with a soft-max output and a log-likelihood loss, as well as the semantic similarity between the triggers and their associated sentences using contrastive loss––we weigh both objectives equally.
Through this joint learning, our trigger representations can capture label knowledge as well as semantic information.
We use these representations to improve model training by generating weakly labeled data via soft matching on the unlabeled sentences. 
More specifically, for each unlabeled sentence, we first calculate the semantic similarity between the sentence and all collected triggers and then filter out all triggers where the similarity distance is larger than our fixed threshold. 
We then generate a trigger-aware sentence encoding for each threshold-passing trigger and feed these encodings into a downstream classifier for label inference. 
Finally, we conduct majority vote over outputted label sequences to finalize our weak labels for the unlabeled sentence. 
In this manner we are able to train over more data, where a good portion of it is weakly labeled.

\subsection{Input: Natural Language}
\label{sec:modelnl}
When natural language is inputted into the system, our module grows training data via soft-matching between logical forms parsed from natural language explanations and unlabeled sentences.
The module follows the Neural Execution Tree framework of ~\citep{qin2020learning} when dealing with natural language.
First, the explanation is parsed into a logical form by a semantic parser.
Previous works have suggested using similar logical forms to improve model training by strict matching on the pool of unlabeled sentences to generate additional labeled data.
However, ~\citep{qin2020learning} proposes an improved model training paradigm, which relaxes this strict matching constraint, subsequently improving weak labeling coverage and allowing for a larger pool of unlabeled data to be used for model training.
Our module does assume each NL explanation can be broken down into a logical form composed of clauses consisting of predicates from four categories––hence the auto-suggest feature in the UI. At weak labeling time, the module scores how likely a given unlabeled sentence fits each clause and then constructs an aggregate score representing the match between the logical form and the unlabeled sentence. If the final score is above configurable thresholds, we weakly label the sentence with the appropriate label.

\begin{figure}[!t]
	\centering
	\includegraphics[width=\linewidth]{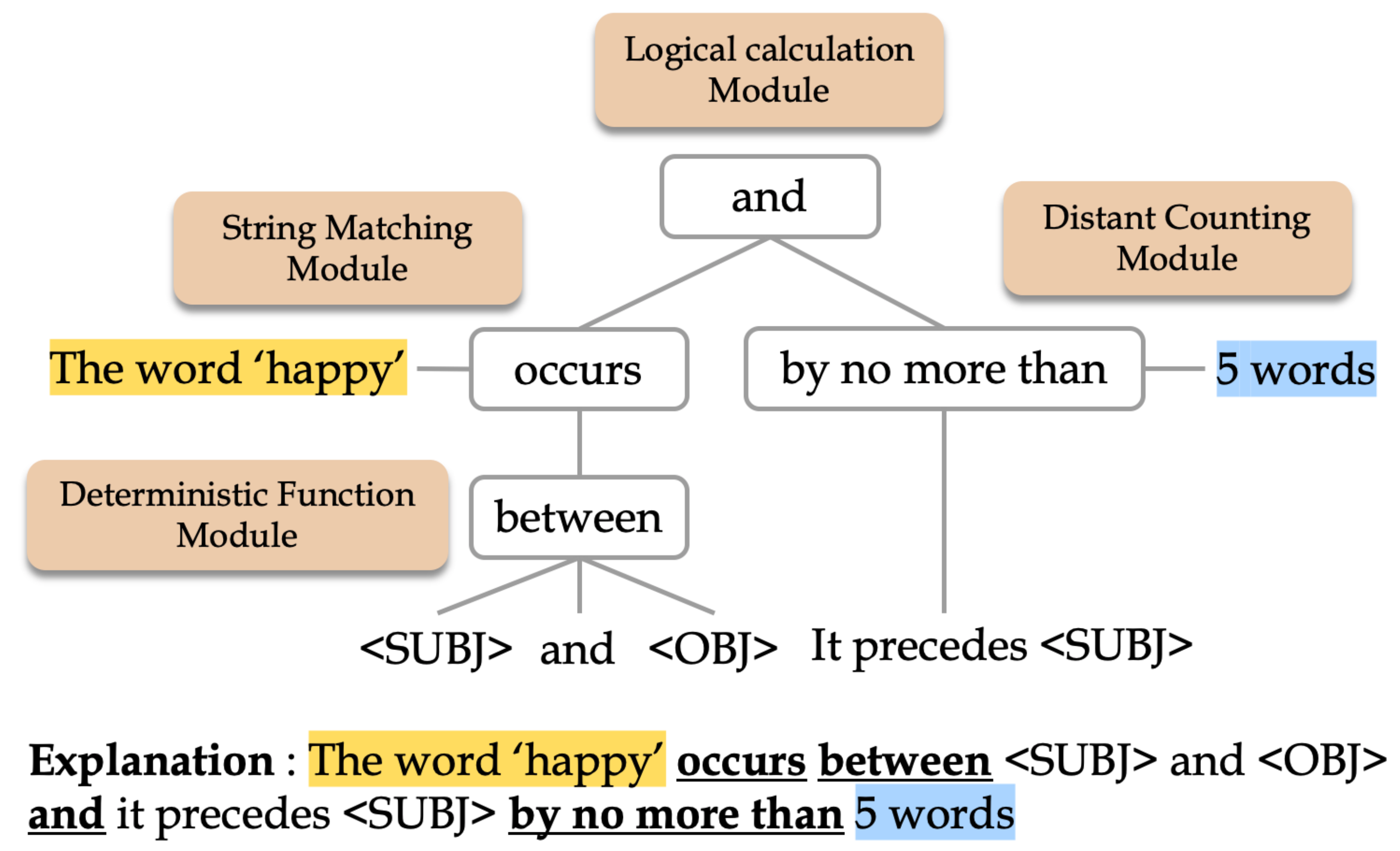}
	\caption{Weakly labeling module for exploiting natural language explanation. the keyword is 'happy'}
	\label{fig:model}
\end{figure}

As shown in ~\figref{fig:model}, the scoring portion of our module has four parts: \textit{String Matching Module}, \textit{Distant Counting Module}, \textit{Deterministic Function Module}, and the \textit{Logical Calculation Module}. 
The first three modules are responsible for evaluating if different clauses in the logical form are applicable for the given unlabeled sentence, while the \textit{Logical Calculation Module's} job is to aggregate scores between the various clauses.
The \textit{String Matching Module} returns a sequence of scores $[s_1, s_2, ..., s_n]$ indicating the similarity between each token $w_i$ and the keyword $q$––``happy" in ~\figref{fig:model}. Our \textit{Distant Counting Module} aims to relax the distance constraint stated in the explanation, ex: ``by no more than 5 words". If the position of keyword $q$ strictly satisfies the constraint, the score is set to 1, otherwise the score decreases as the constraint is less satisfied. Finally, the \textit{Deterministic Function Module} deals with deterministic predicates like ``LEFT", ``BETWEEN", which can only be exactly matched in terms of the keyword $q$. Scores are the aggregated by the \textit{Logical Calculation Module} to output a final relevancy score.
\section{Experiments}
We conduct extensive experiments investigating label efficiency to prove the effectiveness of our annotation models.
We found that using natural language explanations for RE and SA, and trigger explanations for NER provided the best results.
For the downstream model portion of our weak supervision framework, we use common supervised method for each task:
(1-RE) BLSTM+ATT~\cite{bahdanau2014neural} adds an attention layer onto LSTM to encode an sequence.
(2-SA) ATAE-LSTM~\cite{wang2016attention} combines the aspect term information into both the embedding layer and attention layer to help the model concentrate on different parts of a sentence.
(3-NER) BLSTM+CRF~\cite{DBLP:conf/acl/MaH16} encodes character sequences into a vector and concatenates the vector with pre-trained word embeddings to feed into word-level BLSTM. Then, it applies a CRF layer to predict sequence labels. Then we compare these methods as baselines.

\paragraph{Tasks and Datasets} We test our implementation on three tasks: RE, SA, NER.
We use TACRED~\cite{zhang2017position} for RE, Restaurant review from SemEval 2014 Task 4 for SA, and Laptop reviews~\cite{pontiki2016semeval} for NER.

\begin{figure}[t]
\hfill%
    \begin{subfigure}[t]{0.48\linewidth}
    \hspace*{-.2in}
    \centering
    \subcaptionbox{relation extraction}{%
    \includegraphics[width=4.4cm,height=3.9cm]{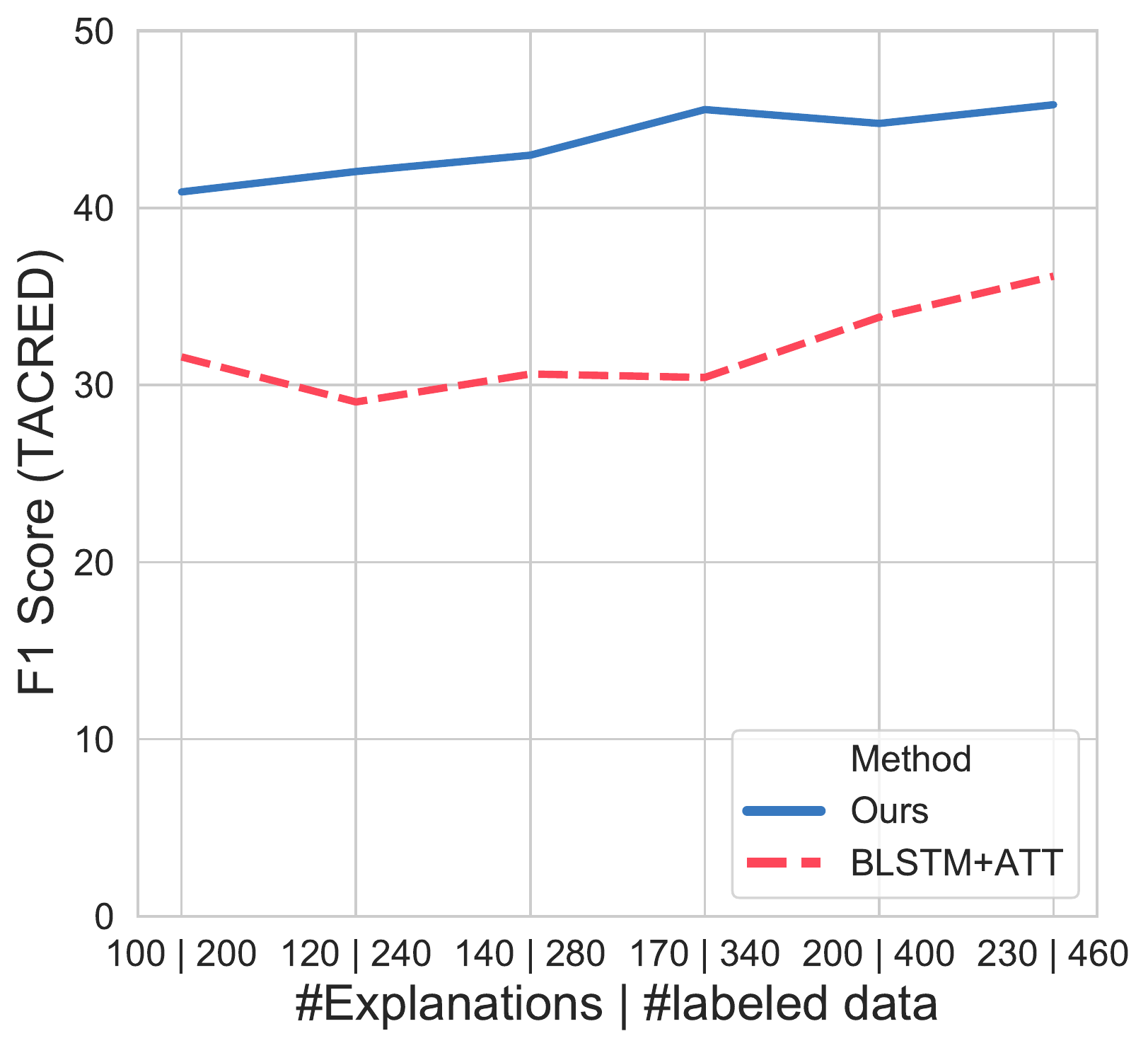}
    }
    \end{subfigure}
\hspace*{\fill}%
    \centering
    \begin{subfigure}[t]{0.48\linewidth}
    \centering
    \subcaptionbox{sentiment analysis}{%
    \includegraphics[width=4.4cm,height=3.9cm]{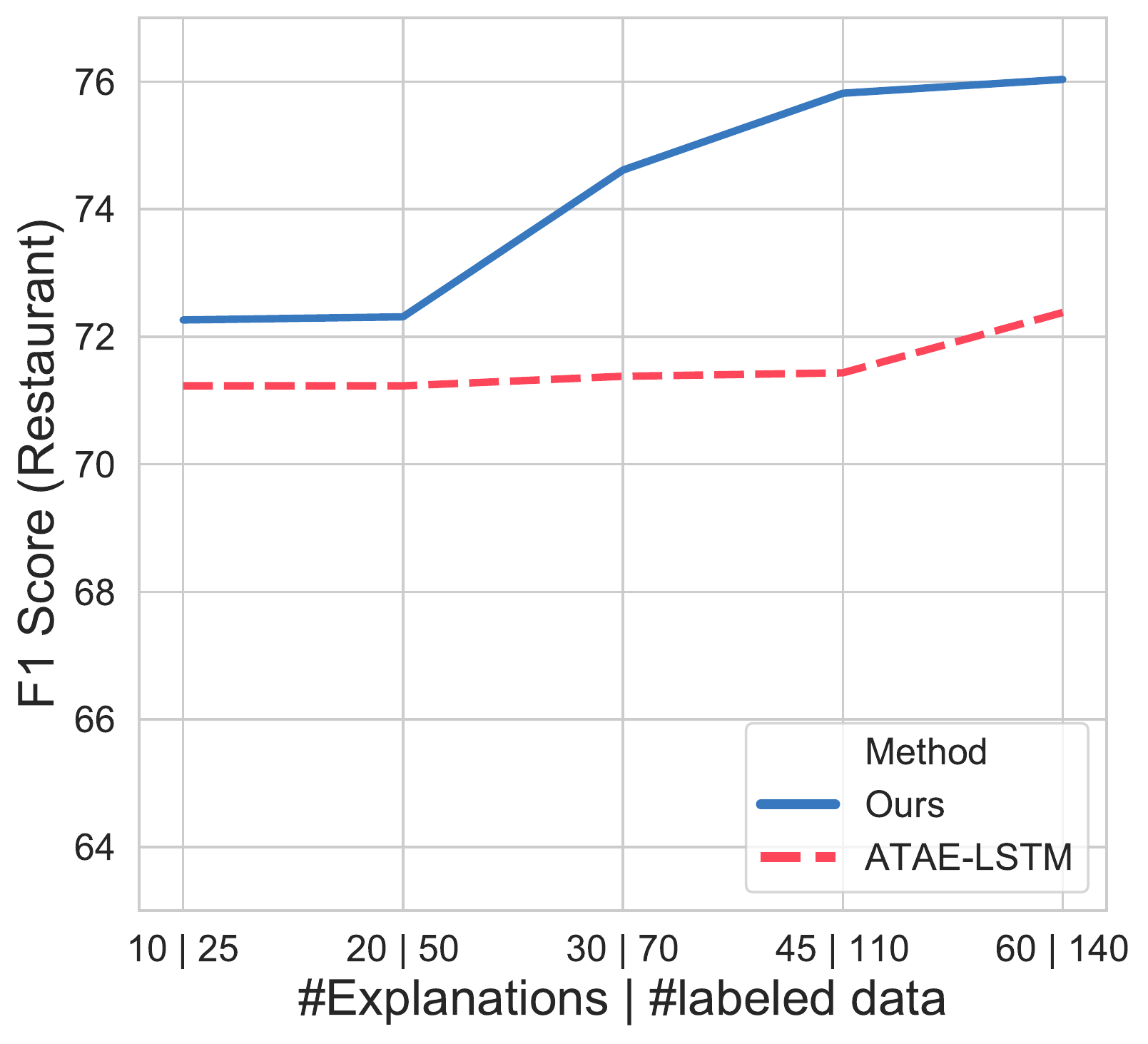}
    }
    \end{subfigure}
    ~ 
    \begin{subfigure}[t]{0.48\linewidth}
    \vspace*{.1in}
    \centering
    \subcaptionbox{named entity recognition}{%
    \includegraphics[width=4.4cm,height=3.9cm]{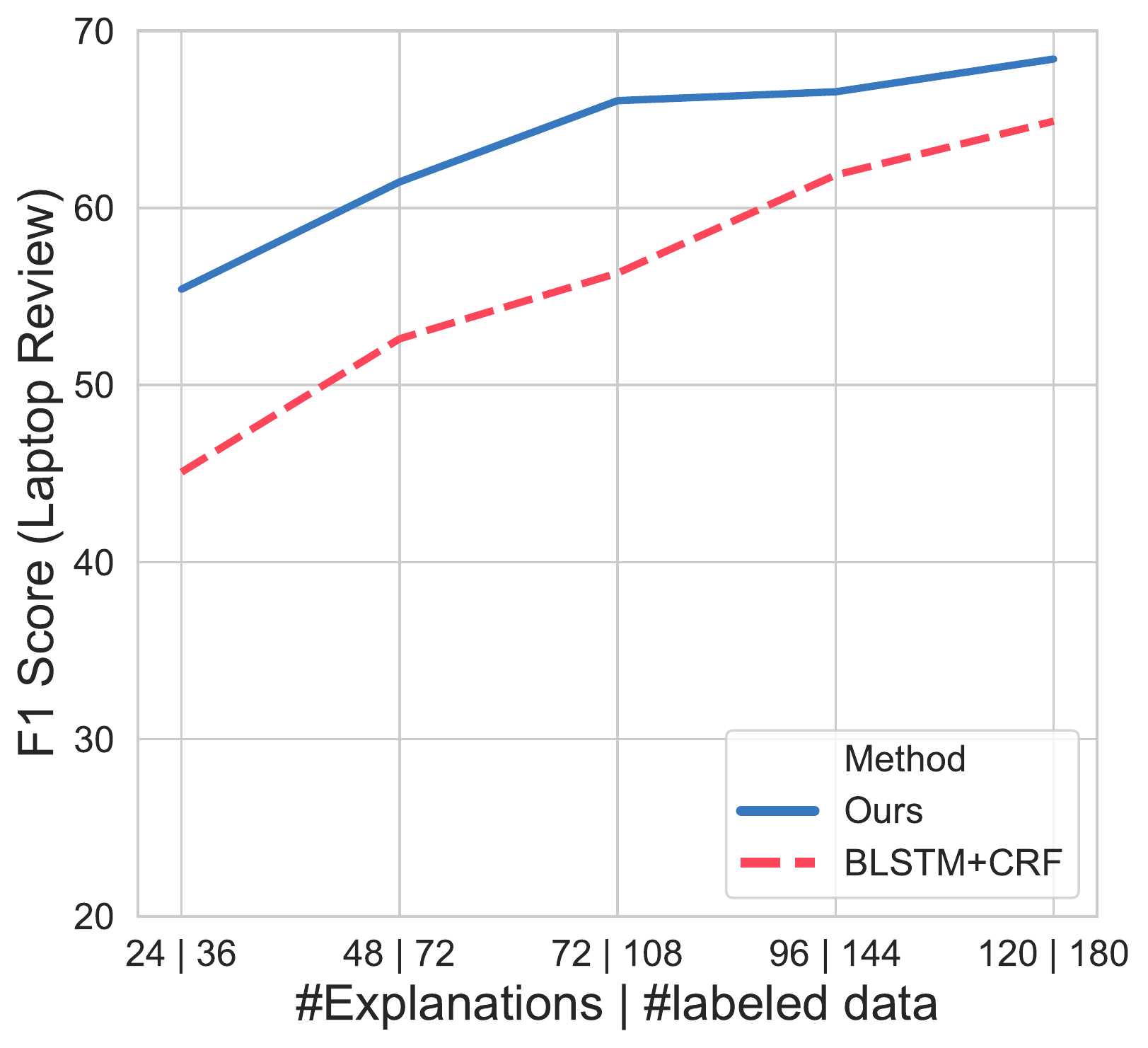}
    }
    \end{subfigure}
\caption{\textbf{Label Efficiency}. We choose commonly-used supervised baselines for comparison. \vspace{-10pt}}
\label{fig:labeleff}
\end{figure}

\paragraph{Label Efficiency}
We claim that when starting with little to no labeled data, it is more effective to ask annotators to provide a label and an explanation for the label, than to just request a label. To support this claim, we conduct experiments to demonstrate the label efficiency of our explanation-leveraging-model.
We found that the time for labeling one instance plus providing an explanation takes 2X times more time than just simply providing a label. 
Given this annotation time observation, we compare the performance between our improved training process and the traditional label-only training process by holding annotation time constant between the two trials. This means we expose the label-only supervised model to the appropriate multiple of labeled instances that the label-and-explanation supervised model is shown ~\figref{fig:labeleff}. Each marker on the x-axis of the plots indicate a certain interval of annotation time, which is represented by the number of label+explanations our augmented model training paradigm is given vs. how many labels the traditional label-only model training is shown. We use the commonly used F-1 metric to compare the performances.
As shown in ~\figref{fig:labeleff}, we see that our model not only is more time and label efficient than the traditional label-only training process, but it also outright outperforms the label-only training process. 
Given these results, we believe it is worth to request a user to provide both a label and an explanation for the label. Not only does the improvement in performance justify the extra time required to provide the explanation, but we also can achieve higher performance with fewer datapoints / less annotation time.
\section{Related Works}
Leveraging natural language explanations for additional supervision has been explored by many works. ~\citep{srivastava2017joint} first demonstrated the idea of using natural language explanations for weak labeling by jointly training a task-specific semantic parser and label classifier to generate weak labels.
This method is limited though, as the parser is too tightly coupled to the already labeled data, thus their weak learning framework is not able to build a much larger dataset than the one it already has.
To address this issue, ~\citep{hancock2018training} proposed a weak supervision framework that utilizes a more practical rule-based semantic parser. The parser constructs a logical form for an explanation that is then used as a labeling function––this resulted in a significant increase of the training set.
Another effort to incorporate explanations can be found in ~\citep{camburu2018snli} work to extend the Stanford Natural Language Inference dataset with natural language explanations––this extension was done for the important textual entailment recognition task. They demonstrate the usefulness of explanations as an additional training signal for learning more comprehensive sentence representations.
Even earlier ~\citep{andreas2016neural} explored breaking down natural language explanation into linguistic sub-structures for learning collections of neural modules which can be assembled into neural networks.
Our framework is very related to the above weak supervision methods via explanation. 

Another approach to weak supervision is attempting to transfer knowledge from a related source to the target domain corpus~\cite{Lin2018NeuralAL,Lan2020}. ~\citeauthor{weaklabel} (2017) attempts to create weakly labeled NER data for a target language via an annotation projection from a comparable corpus. However, their efforts regard unlabeled words as `\textit{O}', and so it cannot deal with incomplete annotations––a feature an annotation framework must handle.
~\citeauthor{autoner} (2018) and ~\citeauthor{yangner} (2018) proposed using a domain-specific dictionary for matching on the unannotated target corpus. Both efforts employ Partial CRFs~\cite{liu2014domain} which assign all possible labels to unlabeled words and maximize the total probability. This approach addresses the incomplete annotation problem, but heavily relies on a domain-specific seed dictionary.
\section{Conclusion}
\label{sec:con}
In this paper, we propose an open-source web-based annotation framework \texttt{LEAN-LIFE} that not only allows an annotator to provide the needed labels for a task, but can also capture explanation for each labeling decision. Such explanations enable a significant improvement in model training while only doubling per instance annotation time. This increase in per instance annotation time is greatly outweighed by the benefits in model training, especially in a low resource settings, as proven by our experiments. This is an important consideration for any annotation framework, as the quicker the framework is able to train annotation recommendation models to reach high performance, the sooner the user receives useful annotation recommendations, which in turn cut down on the annotation time required per instance. 

Better training methods also allow us to fight the potential generation of noisy data due to inaccurate annotation recommendations.
We hope that our work on \texttt{LEAN-LIFE} will allow for researches and practitioners alike to more easily obtain useful labeled datasets and models for the various NLP tasks they face.


\section*{Acknowledgements}
This research is based upon work supported in part by the Office of the Director of National Intelligence (ODNI), Intelligence Advanced Research Projects Activity (IARPA), via Contract No. 2019-19051600007, NSF SMA 18-29268, and Snap research gift. The views and conclusions contained herein are those of the authors and should not be interpreted as necessarily representing the official policies, either expressed or implied, of ODNI, IARPA, or the U.S. Government. We would like to thank all the collaborators in USC INK research lab for their constructive feedback on the work.
\bibliography{acl2020}
\bibliographystyle{acl_natbib}
\end{document}